\documentclass{article}
\usepackage[utf8]{inputenc}
\pdfoutput=1
\usepackage[sc,osf]{mathpazo}
\usepackage[margin=1.1in,letterpaper]{geometry}
\usepackage[sort&compress,numbers]{natbib}
\usepackage[colorlinks=true,citecolor=blue,breaklinks]{hyperref}
\usepackage[hyphenbreaks]{breakurl} %for arXiv's idiotic hyperref!
\usepackage{caption}
\usepackage{subcaption}

\makeatletter
\newlength\aftertitskip     \newlength\beforetitskip
\newlength\interauthorskip  \newlength\aftermaketitskip

\def\maketitle{\par
 \begingroup
   \def\thefootnote{\fnsymbol{footnote}}
   \def\@makefnmark{\hbox to 4pt{$^{\@thefnmark}$\hss}}
   \@maketitle \@thanks
 \endgroup
\setcounter{footnote}{0}
 \let\maketitle\relax \let\@maketitle\relax
 \gdef\@thanks{}\gdef\@author{}\gdef\@title{}\let\thanks\relax}

\def\@startauthor{\noindent \normalsize\bf}
\def\@endauthor{}
\def\@starteditor{\noindent \small {\bf Editor:~}}
\def\@endeditor{\normalsize}
\def\@maketitle{\vbox{\hsize\textwidth
 \linewidth\hsize \vskip \beforetitskip
 {\begin{center} \LARGE\@title \par \end{center}} \vskip \aftertitskip
 {\def\and{\unskip\enspace{\rm and}\enspace}%
  \def\addr{\small\it}%
  \def\email{\hfill\small\tt}%
  \def\name{\normalsize\bf}%
  \def\AND{\@endauthor\rm\hss \vskip \interauthorskip \@startauthor}
  \@startauthor \@author \@endauthor}
}}

\makeatother

\pdfoutput=1                    % force pdflatex to run

\usepackage{floatrow} % For putting table and figure sude by side
\newfloatcommand{capbtabbox}{table}[][\FBwidth]

\usepackage{graphicx}
\usepackage{amsmath,amsthm,amssymb,bm} % define this before the line numbering.
\usepackage{amsfonts}
\usepackage{xspace}
\usepackage{array}
\usepackage{enumerate}
\usepackage{algorithm}
\usepackage{algorithmicx,algpseudocode}
\usepackage{stmaryrd}

\numberwithin{equation}{section}

\newcommand{\reals}{\mathbf{R}}

\newtheorem{theorem}{Theorem}

\newtheorem{claim}{Claim}
\theoremstyle{definition}

\theoremstyle{remark}

\newcommand{\eps}{\varepsilon}

\newcommand{\dpp}{\textsc{Dpp}\xspace}

\newcount\Comments  % 0 suppresses notes to selves in text
\usepackage{color}
\definecolor{darkgreen}{rgb}{0,0.5,0}
\definecolor{purple}{rgb}{1,0,1}
\newcommand{\comm}[2]{\ifnum\Comments=1\textcolor{#1}{#2}\fi}
\title{Fast Sampling for Strongly Rayleigh Measures with Application to Determinantal Point Processes}
\author{\name Chengtao Li \email{ctli@mit.edu}\\
  \name Stefanie Jegelka \email{stefje@csail.mit.edu}\\
  \name Suvrit Sra \email{suvrit@mit.edu}\\
  \addr{Massachusetts Institute of Technology, Cambridge, MA 02139} 
}

\begin{document}
\maketitle

\begin{abstract}
  In this note we consider sampling from (non-homogeneous) strongly Rayleigh probability measures. As an important corollary, we obtain a  fast mixing Markov Chain sampler for Determinantal Point Processes. 
\end{abstract}

\section{Introduction}
Probability distributions over combinatorial families of subsets are important to a variety of problems in machine learning and related areas. Notable examples include discrete probabilistic models \cite{bouchard10,smith08,zhang2015higher,gps89,kulesza2012determinantal} for use in computer vision, computational biology, and Natural Language Processing; combinatorial bandit learning \cite{cesabianchi09}; model compression~\cite{mariet16}; and low-rank matrix approximations \cite{li2016fast}. Consequently, significant recent attention has been paid to sampling rapidly from certain structured discrete  distributions~\cite{gotovos2015sampling,rebeschini2015fast}, as well as from determinantal point processes \cite{anari2016monte,li2016fast,li2016gauss} and progress on sampling by optimization~\cite{ermon2013embed,a2013maddison}.

Amongst these distributions, a widely used class is that of log-submodular measures. Formally, for sets $S,T \subseteq V$, a \emph{log-submodular} measure $\pi: 2^V \to \reals_+$ satisfies the inequality
\begin{equation}
  \label{eq:3}
  \log \pi(S) + \log \pi(T) \geq \log \pi(S \cup T) + \log \pi(S \cap T).
\end{equation}
Log-submodular measures are useful to several applications in machine learning and computer vision \cite{kulesza2012determinantal,barinova12}; more generally, submodular functions are widely important across machine learning~\cite{nipstut,icmltut2,bachtut}.

In this note, we focus on a specific subclass of log-submodular measures, namely, \emph{strongly Rayleigh (SR) measures}. These measures are intimately related to stable polynomials, a viewpoint first established in~\cite{borcea2009negative}, which has proved key to uncovering their remarkable properties, both for modeling as well as for fast sampling. For instance, these measures exhibit \emph{negative association}, a strong, ``robust'' notion of negative dependence (we formally define SR measures in Section~\ref{sec:rayleigh}).

\vskip4pt
\noindent We mention below some important examples of SR measures.\\[4pt]
\noindent\textbf{Determinantal Point Processes.} A Determinantal Point Process (DPP) is a measure over subsets given by the principal minors of a positive semidefinite matrix $K \in \mathbb{R}^{N \times N}$ with eigenvalues in $[0,1]$. Its marginal probabilities satisfy
\begin{equation}
  \label{eq:4}
  \mathrm{Pr}(S \subseteq T) = \det(K_S),
\end{equation}
where $K_S$ is the submatrix indexed by the elements in $S$, and $T$ is the random set distributed as a DPP. DPPs arise in random matrix theory, combinatorics, machine learning, matrix approximations, and many other areas; see e.g.,~\cite{macchi1975,lyons2003,lyons2014,bufetov2012,borodin2009,soshnikov2000,kulesza2012determinantal,hough2005,borodin2000,gorin.lect,li2016gauss}.

\vskip4pt
\noindent\textbf{(Weighted) regular and balanced matroids.} The uniform distribution over the bases of certain matroids (regular matroids and balanced matroids \cite{feder1992balanced,pemantle2014}) is SR, most notably, the uniform distribution over spanning trees in a graph. Here, spanning trees are viewed as subsets of edges, and the distribution is over subsets of edges. 

\vskip4pt
\noindent\textbf{Product measures / Bernoullis conditioned on their sum.} Assume there is a weight $q_i \in [0,1]$ for each element $i \in V$. The product measure $\pi(S) = \prod_{i \in S}q_i \prod_{j \notin S}(1-q_j)$ is SR, as is its conditioning on sets of a specific cardinality $k$, i.e., $\pi'(S) = \pi(S \mid |S|=k)$ or $\pi'(S)=0$ if $|S| \neq k$, and $\pi'(S) \propto \pi(S)$ otherwise. 

Strongly Rayleigh measures have been underlying recent progress in approximation algorithms \cite{gharan11,anari15,deshpande06,li2016fast}, graph sparsification \cite{frieze14,spielman08}, extensions to the Kadison-Singer problem~\citep{anari2014}, finite extensions to free probability~\cite{marcus2016}, and concentration of measure results~\cite{pemantle2014}.

\paragraph{Contributions.}
Despite their importance, efficient sampling methods are only known for special cases of SR measures. In this note, we derive a provably fast mixing Markov Chain for efficiently sampling general SR measures. For our analysis, we use the recent result of \cite{anari2016monte} (that analyzes fast mixing for the subclass of $k$-homogeneous SR measures), along with the closedness properties of SR measures established in the landmark work~\cite{borcea2009negative}.

%\subsection{Markov Chains and Mixing time}

\section{Sampling from Strongly Rayleigh Distributions}
\label{sec:rayleigh}
\emph{Strongly Rayleigh (SR)} distributions capture the strongest form of negative dependence, while enjoying a host of other notable properties~\cite{borcea2009negative}. Several important distributions exhibit the strong Rayleigh property, for example, uniform distributions over spanning trees in graphs, and more generally, the widely occurring \emph{Determinantal Point Processes}. 
A distribution is strongly Rayleigh if its generating polynomial $p_{\pi}: \mathbb{C}^N \to \mathbb{C}$, 
\begin{equation}
  p_{\pi}(z) = \sum_{S \subseteq V} \pi(S) \prod_{i \in S}z_i\label{eq:1}
\end{equation}
is \emph{real stable}. This means that if $\Im(z_i)>0$ for all arguments $z_i$ of $p_{\pi}(z)$, then $p_{\pi}(z) \neq 0$.

\paragraph{Markov Chain Sampling.} 
We sample from $\pi$ via a Markov Chain Monte Carlo method (MCMC), i.e., we run a Markov Chain with state space $2^V$ (the power set of $V$). All the chains discussed here are ergodic. The \emph{mixing time} of the chain indicates the number of iterations $t$ that we must perform (after starting from an arbitrary set $S_0 \in 2^V$) before we can consider $S_t$ a valid sample from $\pi$. Formally, if $\delta_{S_0}(t)$ is the total variation distance between the distribution of $S_t$ and $\pi$ after $t$ steps, then $\tau_{S_0}(\eps) = \min\{t: \delta_{S_0}(t')\le \eps,\ \forall t'\ge t\}$ is the mixing time to sample from a distribution $\epsilon$-close to $\pi$ in terms of total variation distance. We say that the chain mixes fast if $\tau_{S_0}$ is polynomial in $N$.

\paragraph{Existing samplers.}
Efficient sampling techniques have been studied for special cases of SR distributions. A popular method for sampling from Determinantal Point Processes uses the spectrum of the defining kernel \citep{hough2005}. Generic MCMC samplers can also be derived, for example, previous work used a simple add-delete Metropolis-Hasting chain \citep{kang2013fast}. Starting with an arbitrary set $S \subseteq V$, we sample a point $t \in V$ uniformly at random. If $t \in S$, we remove $t$ with probability $\min\{1, \pi(S \setminus \{t\}) / \pi(S)\}$; if $t \notin S$, we add it to $S$ with probability $\min\{1, \pi(S \cup \{t\}) / \pi(S)\}$. Algorithm~\ref{algo:add} shows the (lazy) Markov chain.

\begin{algorithm}[htbp]
  \caption{Add/delete (Metropolis-Hasting) sampler \label{algo:add}}
	\begin{algorithmic} 
	\Require{SR distribution $\pi$}
	\State Initialize $S\subseteq V$
	\While{not mixed}
		\State Let $b=1$ with probability $\tfrac{1}{2}$
		\If{$b=1$}
			\State Pick $t\in V$ uniformly at random
			\If{$t\in S$}
				\State $S = S\backslash \{t\}$ with probability $\min\{1, \pi(S \setminus \{t\}) / \pi(S)\}$
			\Else
				\State $S = S\cup \{t\}$ with probability $\min\{1, \pi(S \cup \{t\}) / \pi(S)\}$
			\EndIf
		\Else
			\State Do nothing
		\EndIf
	\EndWhile
\end{algorithmic}
\end{algorithm}

\begin{algorithm}[htbp]
  \caption{Gibbs exchange sampler \label{algo:exchange}}
	\begin{algorithmic} 
	\Require{Homogeneous SR distribution $\pi$}
	\State Initialize $S\subseteq V$, $\pi(S)>0$
	\While{not mixed}
		\State Let $b=1$ with probability ${1\over 2}$
		\If{$b = 1$}
			\State Pick $s\in S$ and $t\notin S$ uniformly randomly	
			\State $S = S\cup\{t\}\backslash\{s\}$ with probability $\min\{1,\pi(S\cup\{t\} \setminus \{s\}) / \pi(S)\}$	
		\Else
			\State Do nothing
		\EndIf
	\EndWhile
	\end{algorithmic}
\end{algorithm}

The add-delete chain can work well in practice \cite{kang2013fast}, however, it does not always mix fast. An \emph{elementary} Determinantal Point Process has non-zero measure only on sets of a fixed cardinality; for such a process (or a process close to it), the chain will stall or mix slowly.

Another special case of SR distributions are \emph{homogeneous} SR measures. These measures are nonzero only for some sets of a fixed cardinality $k$. Examples include Bernoulli distributions conditioned on cardinality, uniform distributions on the bases of balanced matroids \cite{feder1992balanced}, and $k$-Determinantal Point Processes. A natural MCMC sampler for these processes takes swapping steps: given a current set $S \subseteq V$, it picks, uniformly at random, points $s \in S$ and $t \notin S$, and swaps them with probability $\min\{1,\pi(S\cup\{t\} \setminus \{s\}) / \pi(S)\}$. % \cite{kang2013fast,li2016fast,belabbas2009landmark}. 
Algorithm~\ref{algo:exchange} formalizes this procedure.
Building upon results in \cite{feder1992balanced}, \citet{anari2016monte} recently showed that the mixing time for the swap sampler for homogeneous SR measures is polynomial in $N$, $k$, and $\log(\frac{1}{\epsilon \pi(S_0)})$. These results are restricted to homogeneous SR measures, and do not hold for arbitrary SR measures.

\subsection{A fast mixing chain for general SR measures}
In this note, we define a projection chain that works for arbitrary SR measures, and whose mixing time is polynomial in $N$, $k$, and $\log(\frac{1}{\epsilon \pi(S_0)})$. In particular, we make the results in \cite{feder1992balanced,anari2016monte} accessible to general SR measures by using specific closure properties \cite{borcea2009negative}.

\begin{algorithm}\small
	\caption{\small Markov Chain for Strongly Rayleigh Distribution}\label{algo:rayleigh}
	\begin{algorithmic} 
	\Require{SR distribution $\pi$}
	% \Ensure{$S$ sampled from $\pi$}
	\State Initialize $R_0\subseteq[2N]$ where $|R_0| = N$ and take $S = R_0\cap V$
	\While{not mixed}
		\State draw $q\sim \text{Unif }[0,1]$
		\State draw $t\in V\backslash S$ and $s\in S$ uniformly randomly
		\If{$q\in[0,{(N-|S|)^2\over 2N^2})$} 
			\State $S = S\cup\{t\}$ with probability $\min\{1,{\pi(S\cup\{t\})\over \pi(S)}\times {|S|+1\over N-|S|}\}$ \Comment{Add $t$}
		\ElsIf{$q\in [{(N-|S|)^2\over 2N^2},{N-|S|\over 2N})$}
			\State $S = S\cup\{t\}\backslash\{s\}$ with probability $\min\{1,{\pi(S\cup\{t\}\backslash\{s\})\over \pi(S)}\}$ \Comment{Exchange $s$ with $t$}
		\ElsIf{$q\in [{N-|S|\over 2N}, {|S|^2 + N(N-|S|)\over 2N^2})$}
			\State $S = S\backslash\{s\}$ with probability $\min\{1,{\pi(S\backslash\{s\})\over \pi(S)}\times {|S|\over N-|S|+1}\}$ \Comment{Delete $s$}
		\Else
			\State Do nothing
		\EndIf
	\EndWhile
\end{algorithmic}
\end{algorithm}

The resulting Markov Chain is shown in Algorithm~\ref{algo:rayleigh}. Interestingly, this sampler uses a mixture of add-delete and swap steps. Hence, intuitively, it preserves the good properties of either type of step. In general the sampled sets can have arbitrary cardinality, and hence add-delete steps are needed. If the distribution concentrates on a certain cardinality, the swap steps gain importance.

This intuition is supported by the following theorem.
\begin{theorem}\label{thm:rayleigh}
If $\pi$ is a SR measure, the mixing time $\tau_{S_0}(\eps)$ of the Markov chain in Algorithm~\ref{algo:rayleigh}
% on $\pi_{sh}$ 
is given by
\begin{align}\label{eq:rayleighmix}
\tau_{S_0}(\eps)\le 2 N^2 \left(\log {N\choose |S_0|} + \log (\pi(S_0))^{-1} + \log \eps^{-1}\right).
\end{align}
\end{theorem}
We may choose the initial set % $T$ in Algorithm \ref{algo:rayleigh} 
such that $S_0$ makes the first term in the sum logarithmic in $N$ ($S_0 = R_0 \cap V$ in Algorithm \ref{algo:rayleigh}).

Theorem~\ref{thm:rayleigh} and Algorithm~\ref{algo:rayleigh} make use of the closure of SR measures under  \emph{symmetric homogenization} \cite{borcea2009negative}. The idea underlying this construction is to introduce a ``shadow'' $V'$ of the ground set $V$, and to construct an $N$-homogeneous SR measure $\pi_{sh}$ on this joint ground set $V \cup V'$. Importantly, the marginal distribution on $V$ under this joint measure is exactly $\pi$. The homogeneous measure $\pi_{sh}$ leads to a fast mixing chain that is, however, not practical to implement. Hence, we reduce it to an equivalent, more efficient chain.

\begin{proof}
  First, we construct a symmetric homogenization of $\pi$, a measure $\pi_{sh}$ on $V \cup V'$:
  \begin{align*}
    \pi_{sh}(R) = \left\{\begin{array}{cc}
        \pi(R\cap [N]){N \choose R\cap [N]}^{-1} & \text{if } |R| = N;\\
        0 & \text{otherwise}.
      \end{array}
    \right.
  \end{align*}
  If $\pi$ is SR, so is its symmetric homogenization $\pi_{sh}$. We use this property to derive a fast-mixing chain.
  
  The results in \cite{anari2016monte} show that a Markov Chain with swap steps mixes rapidly for $\pi_{sh}$. Precisely, they show that for any $k$-homogeneous SR distribution on a ground set of size $M$, a Gibbs-exchange sampler has mixing time
\begin{align*}
  \tau_{R_0}(\eps)\le 2k(M-k)(\log\pi_{sh}(R_0)^{-1} + \log\eps^{-1}).
\end{align*}
Here, $M=2N$ and $k=N$, leading to a mixing time of $2N^2(\log\pi_{sh}(T_0)^{-1} + \log\eps^{-1})$, or, equivalently,
\begin{align}\label{eq:rayleighmix}
  \tau_{S_0}(\eps)\le 2 N^2 \left(\log {N\choose |S_0|} + \log (\pi(S_0))^{-1} + \log \eps^{-1}\right).
\end{align}

It remains to show that the chain in Algorithm~\ref{algo:rayleigh} is equivalent to the Gibbs-exchange sampler for $\pi_{sh}$. In fact, one may be tempted to implement the exchange sampler directly. However, it doubles the size of the ground set to $2N$, and always maintains a set of size $N$. If $N$ is large, this can be impractical.

\begin{claim} \label{cor:rayleigh}
  The mixing time of Markov chain in Algorithm~\ref{algo:rayleigh} has the same bound as Eq.~\eqref{eq:rayleighmix}.
\end{claim}
Our exchange sampler maintains a set $R$ of cardinality $|R| = N$. In each iteration, with probability $\frac{1}{2}$, the sampler does nothing, otherwise it proceeds. If it proceeds, it picks $s\in R$ and $t\in [2N]\backslash R$ uniformly at random, and exchanges them with probability 
\begin{equation}
  \label{eq:2}
  \min\left\{1, \frac{\pi_{sh}(R \cup \{t\} \setminus \{s\})}{\pi_{sh}(R)}\right\}.
\end{equation}
If the exchange is accepted, then the new set is $R \cup \{t\} \setminus \{s\}$.

To consider the projection of this chain onto $V$, let $S = R\cap V$, and $T = V\backslash R$. There are in total four possibilities for locations of $s$ and $t$:
\begin{enumerate}
\item With probability ${|S|(N-|S|)\over N^2}$, $s\in S$ and $t\in T$, and we switch assignment of $s$ and $t$ with probability $\min\{1,{\pi_{sh}(R\cup\{t\}\backslash\{s\})\over \pi_{sh}(R)}\} = \min\{1,{\pi(S\cup\{t\}\backslash\{s\})\over \pi(S)}\}$. This is equivalent to switching elements between $S$ and $T$, i.e., an exchange step on $V$.
\item With probability ${|S|(N-|S|)\over N^2}$, we have $s\notin S$ and $t\notin T$. In this case, independent of whether we exchange $s$ and $t$ or not, the set $S = R \cap V$ remains the same. Hence, in this step, $S$ remains unchanged. % and switch with probability $\min\{1,{\pi(S\cup\{t\})\over \pi(S)}\times {|S|+1\over N-|S|}\}$. This is equivalent to doing nothing to $S$;
\item With probability ${|S|^2\over N^2}$, we have $s\in S$ and $t\notin T$, and we switch with probability $\min\{1,{\pi(S\backslash\{s\})\over \pi(S)}\times {|S|\over N-|S|+1}\}$. This is equivalent to deleting element $s$ from $S$.
\item With probability ${(N-|S|)^2\over N^2}$, we have $s\notin S$ and $t\in T$, and switch with probability $\min\{1,{\pi(S\cup\{t\})\over \pi(S)}\times {|S|+1\over N-|S|}\}$. This is equivalent to adding element $t$ to $S$.
\end{enumerate}
Algorithm \ref{algo:rayleigh} performs those steps with exactly the same probabilities; hence, it is a projection of the exchange chain for $\pi_{sh}$ and has the same mixing time.
\end{proof}

\textbf{Remarks.} 
By using the SR property, we obtain a clean bound for fast mixing. In certain cases, the above chain may mix slower in practice than a pure add-delete chain, since it is ``lazier'', i.e., its probability of stalling is higher. However, it is guaranteed to mix well, and, in other cases, can mix much faster than the pure add-delete chain in \cite{kang2013fast,gotovos2015sampling}. We observe both phenomena in our experiments.

\begin{figure}[t]
\centering
	\begin{subfigure}{.3\textwidth}
	\centering
	\includegraphics[width=\textwidth]{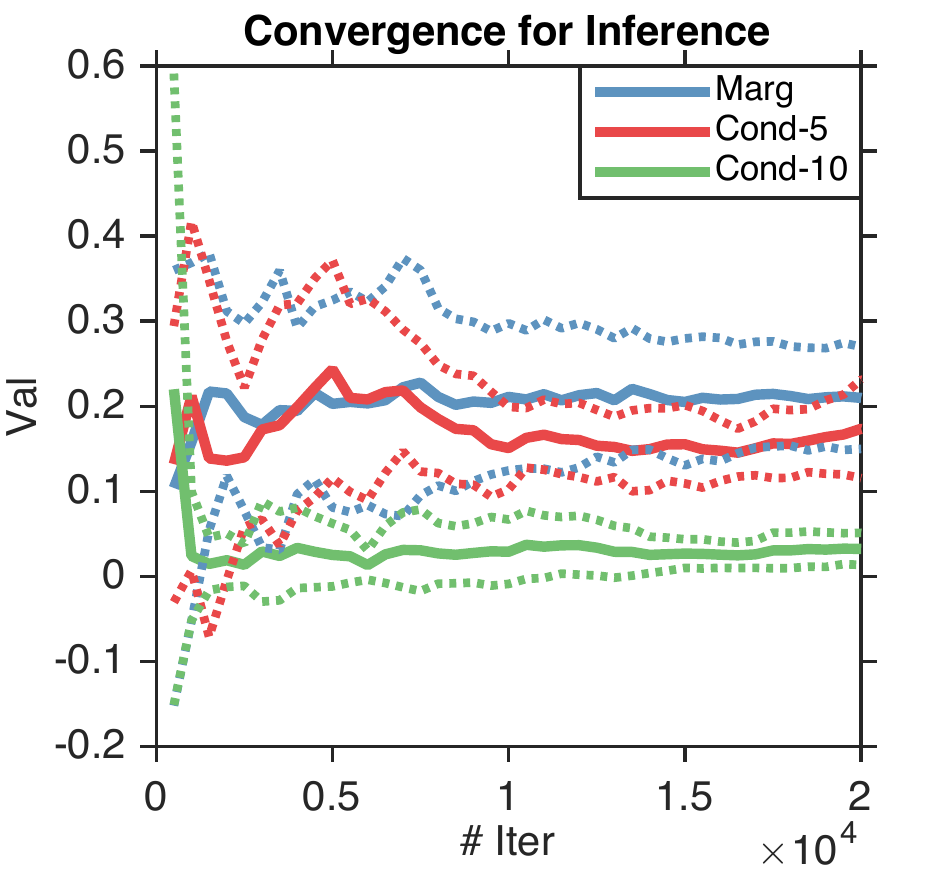}
	\caption{}
	\label{fig:largedpp}
	\end{subfigure}%
	\begin{subfigure}{.32\textwidth}
	\centering
	\includegraphics[width=\textwidth]{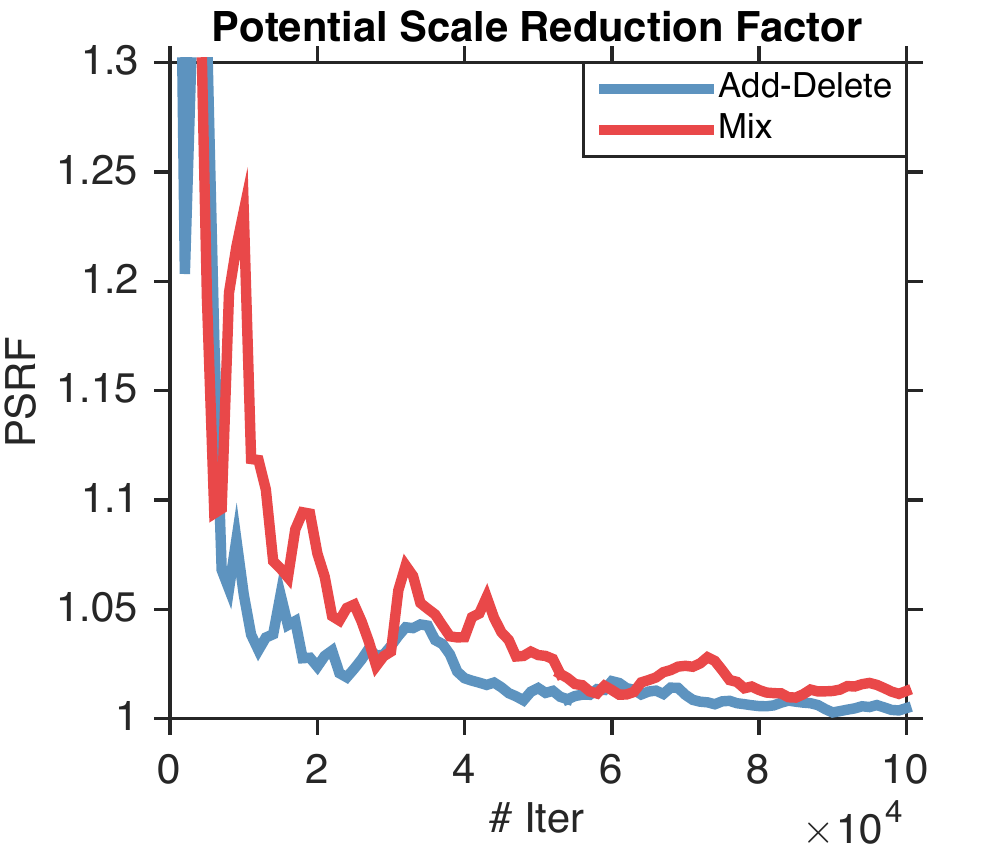}
	\caption{}
	\label{fig:rayleigh1}
	\end{subfigure}%
	\begin{subfigure}{.32\textwidth}
	\centering
	\includegraphics[width=\textwidth]{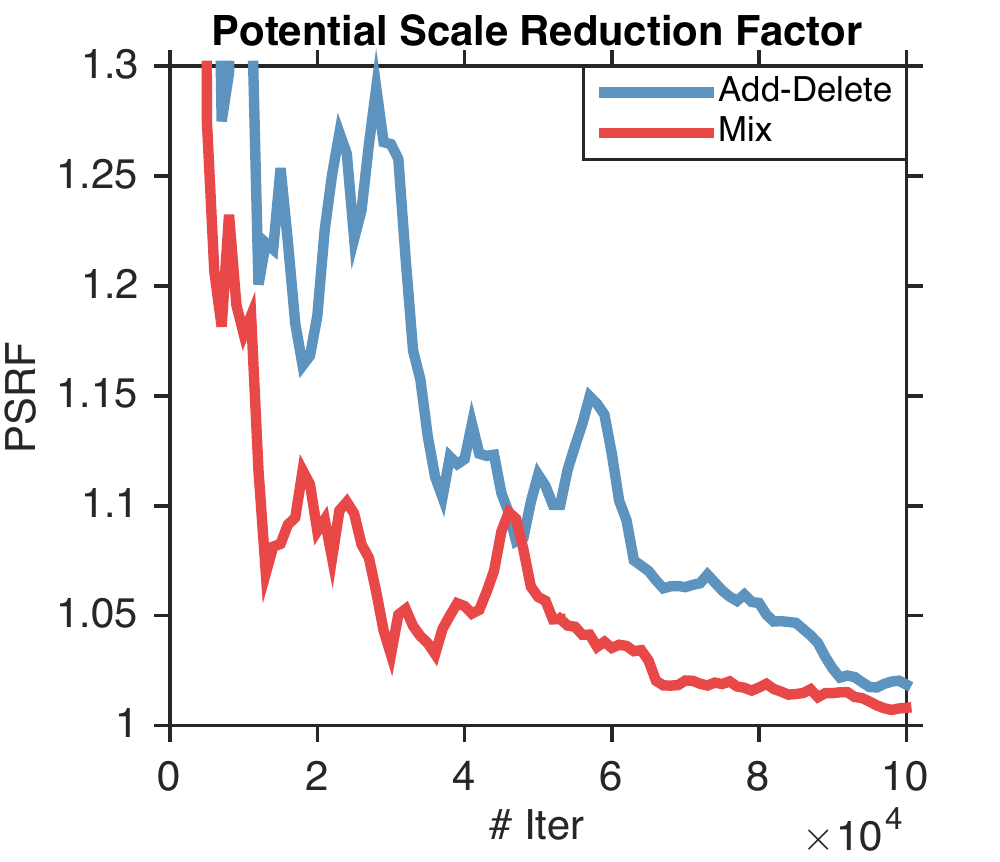}
	\caption{}
	\label{fig:rayleigh2}
	\end{subfigure}
\caption{(a) Convergence of marginal and conditional probabilities by \dpp on uniform matroid, (b,c) comparison between add-delete chain (Algorithm~\ref{algo:add}) and projection chain (Algorithm~\ref{algo:rayleigh}) for two instances: slowly decaying spectrum and sharp step in the spectrum.}
\label{fig:large}
\end{figure}

\section{Experiments}\label{sec:experiments}
Next, we empirically study the how fast our samplers converge. We compare the strongly-Rayleigh chain in Algorithm~\ref{algo:rayleigh}~(\texttt{Mix})  against a simple add-delete chain (\texttt{Add-Delete}). To monitor the convergence of these Markov chains, we use \emph{potential scale reduction factor} (PSRF)~\cite{gelman1992inference,brooks1998general} that runs several chains in parallel and compares within-chain variances to between-chain variances. Typically, PSRF is greater than 1 and will converge to 1 in the limit; if it is close to 1 we empirically conclude that chains have mixed. Throughout experiments we run 10 chains in parallel for estimation, and declare ``convergence'' at a PSRF of 1.05.

We use a \dpp on Ailerons data\footnote{\url{http://www.dcc.fc.up.pt/ 657\~ltorgo/Regression/DataSets.html}} of size 200, and the corresponding PSRF is shown in Fig.~\ref{fig:rayleigh1}. We observe that \texttt{Mix} converges slightly slower than \texttt{Add-Delete} since it is lazier. However, the Add-Delete chain does not always mix fast. Fig.~\ref{fig:rayleigh2} illustrates a different setting, where we modify the eigenspectrum of the kernel matrix: the first 100 eigenvalues are 500 and others 1/500. Such a kernel corresponds to almost an elementary \dpp, where the size of the observed subsets sharply concentrates around 100. Here,  \texttt{Add-Delete} moves very slowly. \texttt{Mix}, in contrast, has the ability to exchange elements and thus converges much faster than \texttt{Add-Delete}.

\bibliographystyle{plainnat}
\setlength{\bibsep}{2pt}
\bibliography{refs_arxiv}

\begin{thebibliography}{40}
\providecommand{\natexlab}[1]{#1}
\providecommand{\url}[1]{\texttt{#1}}
\expandafter\ifx\csname urlstyle\endcsname\relax
  \providecommand{\doi}[1]{doi: #1}\else
  \providecommand{\doi}{doi: \begingroup \urlstyle{rm}\Url}\fi

\bibitem[Anari and Gharan(2015)]{anari15}
Nima Anari and Shayan~O. Gharan.
\newblock Effective-resistance-reducing flows and asymmetric tsp.
\newblock In \emph{IEEE Symposium on Foundations of Computer Science (FOCS)},
  2015.

\bibitem[Anari and Gharan(2014)]{anari2014}
Nima Anari and Shayan~Oveis Gharan.
\newblock The kadison-singer problem for strongly rayleigh measures and
  applications to asymmetric tsp.
\newblock \emph{arXiv preprint arXiv:1412.1143}, 2014.

\bibitem[Anari et~al.(2016)Anari, Gharan, and Rezaei]{anari2016monte}
Nima Anari, Shayan~Oveis Gharan, and Alireza Rezaei.
\newblock {M}onte {C}arlo {M}arkov chain algorithms for sampling strongly
  {R}ayleigh distributions and determinantal point processes.
\newblock \emph{COLT}, 2016.

\bibitem[Bach(2013)]{bachtut}
Francis Bach.
\newblock \emph{Learning with Submodular Functions: A Convex Optimization
  Perspective}.
\newblock Foundations and Trends in Machine Learning, 2013.

\bibitem[Barinova et~al.(2012)Barinova, Lempitsky, and Kohli]{barinova12}
Olga Barinova, Victor Lempitsky, and Pushmeet Kohli.
\newblock On detection of multiple object instances using {H}ough transforms.
\newblock \emph{IEEE Trans. on Pattern Analysis and Machine Intelligence},
  9:\penalty0 1773--1784, 2012.

\bibitem[Bilmes(2013)]{nipstut}
Jeff Bilmes.
\newblock Deep mathematical properties of submodularity with applications to
  machine learning.
\newblock Tutorial at the Conference on Neural Information Processing Systems
  (NIPS), 2013.

\bibitem[Borcea et~al.(2009)Borcea, Br{\"a}nd{\'e}n, and
  Liggett]{borcea2009negative}
Julius Borcea, Petter Br{\"a}nd{\'e}n, and Thomas Liggett.
\newblock Negative dependence and the geometry of polynomials.
\newblock \emph{Journal of the American Mathematical Society}, 22\penalty0
  (2):\penalty0 521--567, 2009.

\bibitem[Borodin(2009)]{borodin2009}
Alexei Borodin.
\newblock Determinantal point processes.
\newblock \emph{arXiv:0911.1153}, 2009.

\bibitem[Borodin and Gorin(2012)]{gorin.lect}
Alexei Borodin and Vadim Gorin.
\newblock Lectures on integrable probability, 2012.

\bibitem[Borodin and Olshanski(2000)]{borodin2000}
Alexei Borodin and Grigori Olshanski.
\newblock Distributions on partitions, point processes, and the hypergeometric
  kernel.
\newblock \emph{Communications in Mathematical Physics}, 211\penalty0
  (2):\penalty0 335--358, 2000.

\bibitem[Bouchard-C\^ot\'e and Jordan(2010)]{bouchard10}
Alexandre Bouchard-C\^ot\'e and Michael~I. Jordan.
\newblock Variational inference over combinatorial spaces.
\newblock In \emph{NIPS}, 2010.

\bibitem[Brooks and Gelman(1998)]{brooks1998general}
Stephen~P Brooks and Andrew Gelman.
\newblock General methods for monitoring convergence of iterative simulations.
\newblock \emph{Journal of computational and graphical statistics}, pages
  434--455, 1998.

\bibitem[Bufetov(2012)]{bufetov2012}
Alexander~I Bufetov.
\newblock Infinite determinantal measures.
\newblock \emph{arXiv:1207.6793}, 2012.

\bibitem[Cesa-Bianchi and Lugosi(2009)]{cesabianchi09}
Nicolo Cesa-Bianchi and Gabor Lugosi.
\newblock Combinatorial bandits.
\newblock In \emph{COLT}, 2009.

\bibitem[Deshpande et~al.(2006)Deshpande, Rademacher, Vempala, and
  Wang]{deshpande06}
Amit Deshpande, Luis Rademacher, Santosh Vempala, and Grant Wang.
\newblock Matrix approximation and projective clustering via volume sampling.
\newblock \emph{Theory of Computing}, 2:\penalty0 225--247, 2006.

\bibitem[Ermon et~al.(2013)Ermon, Gomes, Sabharwal, and Selman]{ermon2013embed}
Stefano Ermon, Carla~P Gomes, Ashish Sabharwal, and Bart Selman.
\newblock Embed and project: Discrete sampling with universal hashing.
\newblock In \emph{NIPS}, pages 2085--2093, 2013.

\bibitem[Feder and Mihail(1992)]{feder1992balanced}
Tom{\'a}s Feder and Milena Mihail.
\newblock Balanced matroids.
\newblock In \emph{STOC}, pages 26--38, 1992.

\bibitem[Frieze et~al.(2014)Frieze, Goyal, Rademacher, and Vempala]{frieze14}
Alan Frieze, Navin Goyal, Luis Rademacher, and Santosh Vempala.
\newblock Expanders via random spanning trees.
\newblock \emph{SIAM Journal on Computing}, 43\penalty0 (2):\penalty0 497--513,
  2014.

\bibitem[Gelman and Rubin(1992)]{gelman1992inference}
Andrew Gelman and Donald~B Rubin.
\newblock Inference from iterative simulation using multiple sequences.
\newblock \emph{Statistical science}, pages 457--472, 1992.

\bibitem[Gharan et~al.(2011)Gharan, Saberi, and Singh]{gharan11}
Shayan~O. Gharan, Amin Saberi, and Mohit Singh.
\newblock A randomized rounding approach to the {T}raveling {S}alesman
  {P}roblem.
\newblock In \emph{IEEE Symposium on Foundations of Computer Science (FOCS)},
  2011.

\bibitem[Gotovos et~al.(2015)Gotovos, Hassani, and Krause]{gotovos2015sampling}
Alkis Gotovos, Hamed Hassani, and Andreas Krause.
\newblock Sampling from probabilistic submodular models.
\newblock In \emph{NIPS}, pages 1936--1944, 2015.

\bibitem[Greig et~al.(1989)Greig, Porteous, and Seheult]{gps89}
Dorothy~M. Greig, Bruce~T. Porteous, and Allan~H. Seheult.
\newblock Exact maximum a posteriori estimation for binary images.
\newblock \emph{Journal of the Royal Statistical Society}, 51\penalty0 (2),
  1989.

\bibitem[Hough et~al.(2006)Hough, Krishnapur, Peres, and Vir{\'a}g]{hough2005}
J~Ben Hough, Manjunath Krishnapur, Yuval Peres, and B{\'a}lint Vir{\'a}g.
\newblock Determinantal processes and independence.
\newblock \emph{Probability Surveys}, 2006.

\bibitem[Kang(2013)]{kang2013fast}
Byungkon Kang.
\newblock Fast determinantal point process sampling with application to
  clustering.
\newblock In \emph{NIPS}, pages 2319--2327, 2013.

\bibitem[Krause and Jegelka(2013)]{icmltut2}
Andreas Krause and Stefanie Jegelka.
\newblock Submodularity in {M}achine {L}earning: New directions.
\newblock Tutorial at the International Conference on Machine Learning (ICML),
  2013.

\bibitem[Kulesza and Taskar(2012)]{kulesza2012determinantal}
Alex Kulesza and Ben Taskar.
\newblock \emph{Determinantal point processes for machine learning}.
\newblock Foundations and Trends in Machine Learning. Now, 2012.

\bibitem[Li et~al.(2016{\natexlab{a}})Li, Jegelka, and Sra]{li2016fast}
Chengtao Li, Stefanie Jegelka, and Suvrit Sra.
\newblock Fast {DPP} sampling for {N}ystr{\"{o}}m with application to kernel
  methods.
\newblock \emph{ICML}, 2016{\natexlab{a}}.

\bibitem[Li et~al.(2016{\natexlab{b}})Li, Sra, and Jegelka]{li2016gauss}
Chengtao Li, Suvrit Sra, and Stefanie Jegelka.
\newblock Gaussian quadrature for matrix inverse forms with applications.
\newblock \emph{ICML}, 2016{\natexlab{b}}.

\bibitem[Lyons(2003)]{lyons2003}
Russell Lyons.
\newblock Determinantal probability measures.
\newblock \emph{Publications Math{\'e}matiques de l'Institut des Hautes
  {\'E}tudes Scientifiques}, 98:\penalty0 167--212, 2003.

\bibitem[Lyons(2014)]{lyons2014}
Russell Lyons.
\newblock Determinantal probability: basic properties and conjectures.
\newblock \emph{arXiv:1406.2707}, 2014.

\bibitem[Macchi(1975)]{macchi1975}
Odile Macchi.
\newblock The coincidence approach to stochastic point processes.
\newblock \emph{Advances in Applied Probability}, pages 83--122, 1975.

\bibitem[Maddison et~al.(2014)Maddison, Tarlow, and Minka]{a2013maddison}
Chris~J Maddison, Daniel Tarlow, and Tom Minka.
\newblock A* sampling.
\newblock In \emph{NIPS}, 2014.

\bibitem[Marcus(2016)]{marcus2016}
Adam~W Marcus.
\newblock Polynomial convolutions and (finite) free probability, 2016.

\bibitem[Mariet and Sra(2016)]{mariet16}
Zelda Mariet and Suvrit Sra.
\newblock Diversity networks.
\newblock In \emph{ICLR}, 2016.

\bibitem[Pemantle and Peres(2014)]{pemantle2014}
Robin Pemantle and Yuval Peres.
\newblock Concentration of lipschitz functionals of determinantal and other
  strong rayleigh measures.
\newblock \emph{Combinatorics, Probability and Computing}, 23\penalty0
  (01):\penalty0 140--160, 2014.

\bibitem[Rebeschini and Karbasi(2015)]{rebeschini2015fast}
Patrick Rebeschini and Amin Karbasi.
\newblock Fast mixing for discrete point processes.
\newblock \emph{COLT}, 2015.

\bibitem[Smith and Eisner(2008)]{smith08}
David~A. Smith and Jason Eisner.
\newblock Dependency parsing by belief propagation.
\newblock In \emph{EMNLP}, 2008.

\bibitem[Soshnikov(2000)]{soshnikov2000}
Alexander Soshnikov.
\newblock Determinantal random point fields.
\newblock \emph{Russian Mathematical Surveys}, 55\penalty0 (5):\penalty0
  923--975, 2000.

\bibitem[Spielman and Srivastava(2008)]{spielman08}
Dan Spielman and Nikhil Srivastava.
\newblock Graph sparsification by effective resistances.
\newblock In \emph{Symposium on Theory of Computing (STOC)}, 2008.

\bibitem[Zhang et~al.(2015)Zhang, Djolonga, and Krause]{zhang2015higher}
Jian Zhang, Josip Djolonga, and Andreas Krause.
\newblock Higher-order inference for multi-class log-supermodular models.
\newblock In \emph{ICCV}, pages 1859--1867, 2015.

\end{thebibliography}

\end{document}